\pdfoutput=1

\documentclass[11pt]{article}

\usepackage{acl}

\usepackage{times}
\usepackage{latexsym}
\usepackage{graphicx}
\usepackage{booktabs}
\usepackage{multirow}
\usepackage{arydshln}
\usepackage{comment}
\usepackage{booktabs}
\usepackage{amsmath}
\usepackage[T1]{fontenc}

\usepackage[utf8]{inputenc}

\usepackage{microtype}

\usepackage{todonotes}

\def\htc#1{ {\hskip#1\tabcolsep} }

\title{Dive into the Chasm:\\Probing the Gap between In- and Cross-Topic Generalization}

\author{Andreas Waldis\thanks{* Corresponding author andreas.waldis@live.com} $^{1,2}$, Yufang Hou$^{3,1}$, Iryna Gurevych$^{1}$ \\
$^1$Ubiquitous Knowledge Processing Lab (UKP Lab) \\
Department of Computer Science and Hessian Center for AI (hessian.AI)\\
Technical University of Darmstadt\\
 $^2$Information Systems Research Lab, Lucerne University of Applied Sciences and Arts \\
 $^3$IBM Research Europe, Ireland \\
\texttt{\href{http://www.ukp.tu-darmstadt.de/}{www.ukp.tu-darmstadt.de}} \hspace{0.5em} \texttt{\href{http://www.hslu.ch/}{www.hslu.ch}}\\
}
\begin{document}
\maketitle

\begin{abstract}
Pre-trained language models (LMs) perform well in In-Topic setups, where training and testing data come from the same topics. 
However, they face challenges in Cross-Topic scenarios where testing data is derived from distinct topics - such as \textit{Gun Control}. 
This study analyzes various LMs with three probing-based experiments to shed light on the reasons behind the In- vs. Cross-Topic generalization gap.
Thereby, we demonstrate, for the first time, that generalization gaps and the robustness of the embedding space vary significantly across LMs. 
Additionally, we assess larger LMs and underscore the relevance of our analysis for recent models.
Overall, diverse pre-training objectives, architectural regularization, or data deduplication contribute to more robust LMs and diminish generalization gaps. 
Our research contributes to a deeper understanding and comparison of language models across different generalization scenarios.
\footnote{We provide data and code at \href{https://github.com/UKPLab/eacl2024-cross-topic-probing}{https://github.com/UKPLab/eacl2024-cross-topic-probing}.}
\end{abstract}

\newcommand\ALBERT{ALBERT}
\newcommand\BERT{BERT}
\newcommand\BART{BART}
\newcommand\ELECTRA{ELECTRA}
\newcommand\DEBERTA{DeBERTa}
\newcommand\ROBERTA{RoBERTa}
\newcommand\GPT{GPT-2}
\newcommand\TfivethreeB{T5 (3B)}
\newcommand\FLANTfivethreeB{FLAN-T5 (3B)}
\newcommand\GPTNeo{GPT-Neo (2.7B)}
\newcommand\GLOVE{GloVe}

\renewcommand{\sectionautorefname}{§}
\renewcommand{\subsectionautorefname}{§}

\section{Introduction}
\label{sec:introduction}


Probing \citep{belinkov-etal-2017-neural, conneau-etal-2018-cram} is widely used to analyze pre-trained language models (LMs) \citep{devlin-etal-2019-bert, Liu2019RoBERTaAR, he2021deberta, radford2019language}.
It enables a better understanding of how LMs encode information and how it evolves in the architecture by studying linguistic properties such as part-of-speech or dependency-tree parsing \citep{tenney-etal-2019-bert, DBLP:conf/iclr/TenneyXCWPMKDBD19}. 
However, probing methods \citep{Hewitt2019DesigningAI,Hewitt2019ASP,Voita2020InformationTheoreticPW,elazar-etal-2021-amnesic} mainly rely on the general In-Distribution (ID) scenario, where we distribute train and test instances independent and identically.
As a result, other more realistic Out-of-Distribution (OOD) scenarios \citep{Shen2021TowardsOG}, like generalizations regarding forthcoming topics or temporal changes in the language, remain underexplored by probing.

\begin{figure}[t]
  \centering
  \includegraphics[width=0.48\textwidth]{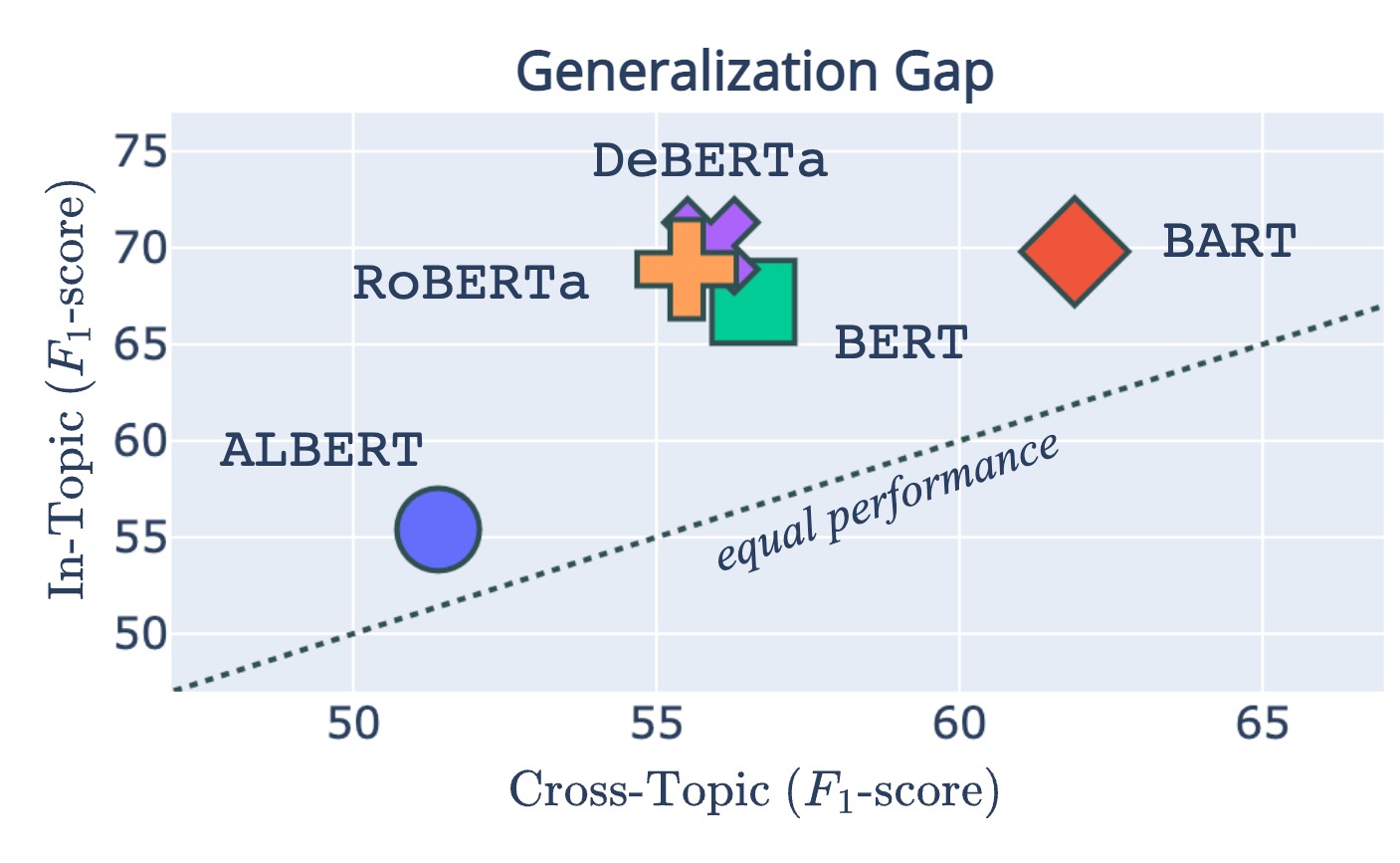}
  \caption{
  Generalization gap of fine-tuning LMs on argumentative \textit{stance detection} \citep{stab-etal-2018-cross} in the In- or Cross-Topic evaluation setup. The dashed line marks the ideal case of equal performance.
  }
  \label{fig:overview}
\end{figure}

Addressing this research gap, we propose - for the first time - a probing-based approach to comprehensively analyze LMs in a challenging OOD setup.
More precisely, we rely on Cross-Topic\footnote{Also known as Cross-Target in \textit{Stance Detection} research.} evaluation where we deliberately withhold instances from specific topics for testing. 
Following \cite{habernal-gurevych-2016-argument,stab-etal-2018-cross}, we define \textit{topic} as the query used to compose a specific dataset - such as arguments covering \textit{gun control} or \textit{marijuana legalization}.
This evaluation setup is highly relevant for challenging Argument Mining (AM) downstream tasks \citep{slonim2021autonomous}.
It allows for simulating, in a controlled setup, how well LMs handle topic-shifts when unseen semantic features (such as topic-specific vocabulary) arise in future and new topics. 
Previous studies found that Cross-Topic argument mining is challenging compared to the In-Topic setup \citep{stab-etal-2018-cross,waldis2023bridging}. 
The major reason lies in the apparent generalization gaps between randomly composing training and testing data (In-Topic) and using distinct groups of topics for training and testing (Cross-Topic).
\autoref{fig:overview} shows such performance gap when fine-tuning on the \textit{UKP ArgMin} dataset \citep{stab-etal-2018-cross} - labeling arguments as in favor, against, or neutral to one of eight topics. 
Notably, we observe gaps between In- and Cross-Topic varying considerably across LMs - with \BART~outperforming the others in the Cross-Topic setup.

Such inconsistencies underline the need to investigate such crucial generalization capabilities.
Thus, we propose extensive probing-based experiments to examine the gap between In- and Cross-Topic generalization and show that embedding spaces of LMs vary considerably regarding their generalizability and robustness. 
In detail, we propose three probing-based experiments to answer the following research questions, considering three linguistic probes (dependency-tree parsing, part-of-speech tagging, and named-entity recognition) based on \textit{UKP ArgMin} dataset:

\paragraph{\textit{How do generalization gaps of LMs differ after pre-training? (\autoref{sec:experiment-1})}}
We find generalization gaps substantially differ across LMs while becoming more prominent for tasks with more semantically difficulties, such as NER. 
In addition, we crucially observe that probing generally underperforms on lexical unseen instances (like highly rare entities), and deduplicating pre-training data provides more robust embedding space when evaluating larger and more recent LMs. 


\paragraph{\textit{How do LMs depend on topic-specific vocabulary? (\autoref{sec:experiment-2})}}
Next, we assess the influence of topic-specific tokens by removing them using amnesic probing and LMs significantly differing in their reliance on and robustness concerning such semantic features.
Interestingly, pre-training objectives or architectural regularization influence robustness, suggesting their potential importance in building robust and competitive LMs.

\paragraph{\textit{How do generalization gaps evolve during fine-tuning? (\autoref{sec:experiment-3})}}
Finally, we re-probe tuned LMs on the \textit{UKP ArgMin} dataset and find that In-Topic fine-tuning erases more linguistic properties than Cross-Topic fine-tuning.

\paragraph{} To sum up, we expand the probing scope to Cross-Topic generalization and highlight probing as a universal tool complementing the study of language models beyond general evaluation setups.
While we focus on an in-depth analysis of In- vs. Cross-Topic generalization gaps, our experimental setup generalizes to other types of OOD scenarios where one verifies generalization regarding other text genres (like the \textit{social media} domain), languages, or temporal changes in the languages \citep{conneau-etal-2018-xnli,Hardalov2021CrossDomainLS,rottger-pierrehumbert-2021-temporal-adaptation,yang-etal-2023-glue}.

\section{In- and Cross-Topic Probing}
\label{sec:topic-information}

The following section formally outlines the probing setup and tasks before elaborating on the generalization gap and comparing the evaluation of In- and Cross-Topic probing.

\subsection{Probing Setup and Tasks}\label{subsec:probe:tasks}
We define a probe $f_p$ comprised of a frozen encoder $h$ and linear classifier $c$ without any intermediate layer. 
This classifier is trained to map instances $X=\{x_1, \ldots, x_n\}$ to targets $Y=\{y_1,  \ldots, y_n\}$ for a given probing task.
Using a frozen LM as $h$, the probe converts $x_i$ into a vector $h_i$. 
In detail, we encode the entire sentence, which wraps $x_i$, and average relevant positions of $x_i$ to find $h_i$.
Relevant positions for the considered probing task are either single tokens for \emph{part-of-speech tagging (POS)}), a span for \emph{named entity recognition (NER)}, or the concatenation of two tokens for \emph{dependency tree parsing (DEP)}.
Then, the classifier $c$ utilizes $h_i$ to generate a prediction $\hat{y}_i$, as shown in \autoref{eqn:prediction}.

\begin{equation}\label{eqn:prediction}
\hat{y}_i = f_p(x_i) = c(h(x_i))
\end{equation}

\subsection{Generalization Gap}

\begin{figure}[t]
  \centering
  \includegraphics[width=0.48\textwidth]{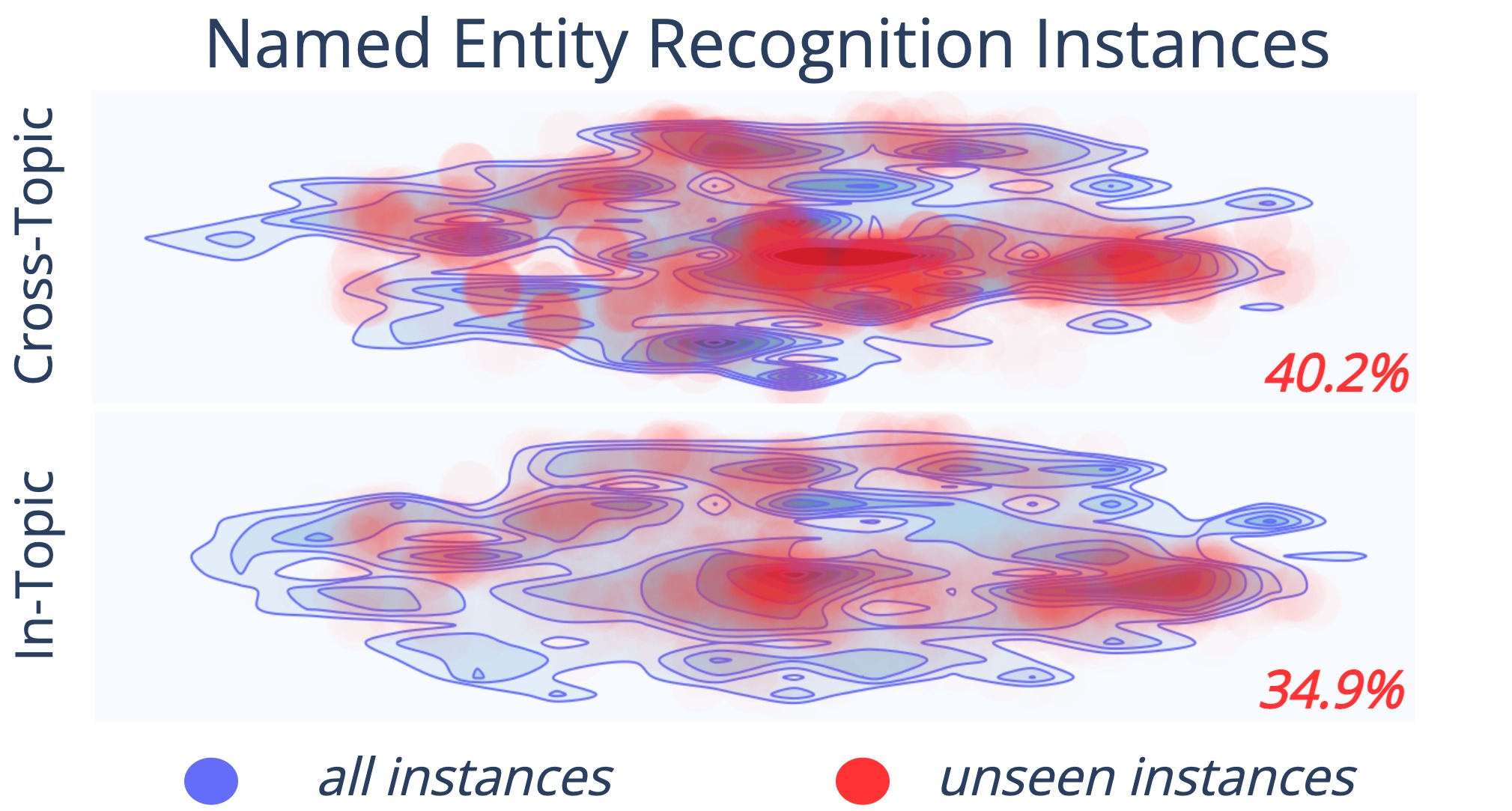}
  \caption{
  Density plot of In- and Cross-Topic NER test instances (blue), encoded with \textit{bert-base-uncased} and reduced with the same t-SNE model \citep{Maaten2008VisualizingDU}.
  While the number of instances is the same, Cross-Topic embodies, with 40.2\%, more \textit{unseen} instances than In-Topic (34.9\%). 
  }
  \label{fig:overview-instances}
\end{figure}

Generalization gaps arise when comparing evaluation setups focusing on different capabilities for the same task. 
This work focuses on gaps in using data from the same (In-Topic) or different topics (Cross-Topic) for training and testing.
We define such topics $T=\{t_1, \ldots, t_m\}$ as the query to collect instances and thereby given by specific datasets \cite{habernal-gurevych-2016-argument,stab-etal-2018-cross} - such as arguments covering \textit{gun control} or \textit{marijuana legalization}.
The In- vs. Cross-Topic gap is visible in \autoref{fig:overview-instances}, which shows how NER instances (in blue) are distributed in the semantic space.
For Cross-Topic, entities cover only specific topics and thereby are less broadly spread, while In-Topic ones are spread more broadly since they cover all datasets' topics.
Simultaneously, we note more lexically \textit{unseen} entities (in red) during training for Cross-Topic.  
Ideally, generalization gaps do not exist since pre-trained language models (LMs) overcome such distribution shifts between different evaluation setups.
However, practically, these gaps vary for different models (\autoref{fig:overview}).

\subsection{Difference between In- and Cross-Topic Evaluation}

By evaluating probing tasks for In- and Cross-Topic, we examine the varying generalization gaps between these setups across different LMs.

\paragraph{Cross-Topic}
With Cross-Topic evaluation, we investigate how well a probe generalizes when the train, dev, and test instances cover distinct sets of topics $\{T^{(train)}, T^{(dev)}, T^{(test)}\}$. 
A probe $f_p$ must generalize across the distribution shift in this setup.
This shift originates because distinct topics cover different specific vocabulary $Z$ - i.e., $Z_{(test)}$ for topics in $T^{(test)}$.
We formally describe this shift, denoted as $\Delta Z$, as the relative complement between topic-specific vocabulary from train and test instances - $\Delta Z = Z_{(train)}\setminus Z_{(test)}$. 
For Cross-Topic, we expect $\Delta Z$ to be large (\autoref{fig:overview-instances}).

\paragraph{In-Topic}
In contrast, $\Delta Z$ is smaller for the In-Topic setup because instances from every split (train/dev/test) cover the same topics. 
We expect similar topic distribution and minor semantic differences within these splits compared to Cross-Topic (\autoref{fig:overview-instances}). 
Thus, we see fewer difficulties for In-Topic because a classifier does not need to generalize across a large distribution shift $\Delta Z$.

\paragraph{Topic-Specific Vocabulary}
\label{para:token-level-topic-information}
As discussed previously, we see topic-specific vocabulary as one main reason for generalization gaps between In- and Cross-Topic because $\Delta Z$ differs for these setups considering a dataset $d$ covering topics $T={t_1, \ldots, t_m}$. 
The topic-specificity of a token $z_i$ is a latently encoded property within the encodings $h_i$ for a token $w_i$. 
To capture this property on the token level, we adopt the approach of \newcite{kawintiranon-singh-2021-knowledge} and use the maximum log-odds-ratio $r_i$ of a token regarding a set of topics $T$. 
Firstly, we calculate the odds of finding the token $w_i$ in a topic $t_j$ as $o_{(w_i, t_j)}=\frac{n(w_i, t_j)}{n(\neg w_i, t_j)}$, where $n(w_i, t_j)$ is the number of occurrences of $w_i$ in $t_j$, and $n(\neg w_i, t_j)$ is the number of occurrences of every other token $\neg w_i$ in $t_j$. 
We then compute $r$ as the maximum log-odds ratio of $w_i$ for all topics in $T$ as $r_{(w_i, T)} = \mathop{max}_{t_j \in T}( log(\frac{o{(w_i, t_j)}}{o_{(w_i, \neg t_j)}}))$.


\section{Experimental Setup}
\label{sec:experimental-design}

We propose three experiments to analyze the varying generalization gap between LMs after pre-training (\autoref{sec:experiment-1}), their dependence on topic-specific vocabulary (\autoref{sec:experiment-2}), and the evolution of these gaps during fine-tuning (\autoref{sec:experiment-3}).
We outline general details about these experiments, while details and results are provided in the subsequent sections.

\paragraph{Models}
We examine how various LMs (\autoref{tab:models}) with varying pre-training objectives or architectural designs differ regarding our probing tasks.
We cover LMs pre-trained using masked language modeling (MLM), next sentence prediction (NSP), sentence order prediction (SOP), language modeling (LM), discriminator (DISC), and denoising autoencoder (DAE) objectives.
As in previous work \citep{Koto2021DiscoursePO}, we group them into the ones pre-trained using token- (MLM) and sentence-objectives (NSP, SOP, or DAE) and four purely token-based pre-trained (MLM, LM, DISC). 
We consider the base-sized variations to compare their specialties in a controlled setup.
Apart from these seven contextualized LMs, we use a static LM with \texttt{GloVe} \citep{pennington-etal-2014-glove}.

\begin{table}[t]
\centering
    \setlength{\tabcolsep}{3pt}
    \resizebox{0.48\textwidth}{!}{%
    \begin{tabular}{lccc}
    \toprule
    \bf Model & \bf \# Params & \bf Objectives & \bf Data \\ \midrule
    \texttt{ALBERT} \citep{Lan2020ALBERTAL} & 12M & MLM + SOP &  16GB  \\
    \texttt{BART} \citep{lewis-etal-2020-bart} & 121M & DAE &  160GB  \\
    \texttt{BERT} \citep{devlin-etal-2019-bert} & 110M & MLM + NSP &  16GB  \\ \midrule
    \texttt{DeBERTa} \citep{he2021deberta} & 100M & MLM &  80GB  \\
    \texttt{RoBERTa} \citep{Liu2019RoBERTaAR} & 110M & MLM &  160GB  \\
    \texttt{ELECTRA} \citep{clark2020electra} & 110M & MLM+DISC &  16GB  \\
    \texttt{GPT-2} \citep{radford2019language} & 117M & LM &  40GB  \\
    \bottomrule
    \end{tabular}
    }
    \caption{Overview of the used LMs trained on MLM, LM, DISC, NSP, SOP, or DAE objectives.}
    \label{tab:models}
\end{table}

\paragraph{Data}
We require a dataset with distinguishable topic annotations to evaluate probing tasks in the In- and Cross-Topic evaluation setup. 
Therefore, we mainly\footnote{We verified our findings with another dataset in the Appendix \autoref{sec:appendix-generalization}.} rely on the \textit{UKP ArgMin} dataset \citep{stab-etal-2018-cross}, which provides 25,492 arguments annotated for their argumentative stance (\textit{pro}, \textit{con}, or \textit{neutral}) towards one of eight distinct topics like \textit{Nuclear Energy} or \textit{Gun Control}.
Using these instances, we heuristically generate at most 40,000 instances for the three linguistic properties \emph{dependency tree parsing} (\textbf{DEP}), \emph{part-of-speech tagging} (\textbf{POS}), or \emph{named entity recognition} (\textbf{NER}) using spaCy.\footnote{We show in the Appendix (\autoref{sec:appendix-label-verification}) that the heuristically generated labels are reliable, and our results are well aligned with previous work. }
Additionally, we consider the main task of the \textit{UKP ArgMin} dataset \citep{stab-etal-2018-cross} -  \emph{argumentative stance detection} (\textbf{Stance}).
Therefore, we have a topic-dependent reference probe to relate the results of other probes and evaluate the generalization ability of LMs on real-world tasks after pre-training.
We use a three-folded setup for all these four probing tasks to consider the full data variability for both In- and Cross-Topic evaluation.
Details about the compositions of these folds and how we ensure a fair comparison between In- and Cross-Topic are provided in the Appendix (\autoref{subsec:appendix:folds}) as well as examples for probing tasks (Appendix \autoref{subsec:appendix:tasks}).

\paragraph{Evaluation}
We primarily report the macro $F_1$ score averaged over the results of evaluating each of the three folds three times using different random seeds. 
Following recent work \citep{voita-titov-2020-information,pimentel-etal-2020-information}, we additionally report information compression $I$ \cite{voita-titov-2020-information} for a holistic evaluation. 
It measures the effectiveness of a probe as the ratio ($\frac{u}{mdl}$) between uniform code length $u=n*log_2(K)$ and minimum description length $mdl$, where $u$ denotes how many bits are needed to encode $n$ instances with label space of $K$.
We follow \textit{online} variation of $mdl$ and use the same ten-time steps $t_{1:11} = \{\frac{1}{1024}, \frac{1}{512}, ..., \frac{1}{2}\}$, where we train a probe for every $t_j$ with a fraction of instances and evaluate with the same fraction of non-overlapping instances.
Exemplary, for, $t_{9}$ we use the first fraction of $\frac{1}{4}$ instances to train and another fraction of $\frac{1}{4}$ to evaluate.
We find the final $mdl$ as the sum of the evaluation losses of every time step $t_{1:11}$. 
For Cross-Topic, we group training instances into two groups of distinct topics and sample the same fraction of instances to train and evaluate. 
Thus, we ensure a similar distribution shift between training and evaluation fractions as in all instances.


\section{The Generalization Gap of LMs}
\label{sec:experiment-1}

\begin{figure}[]
  \centering
  \includegraphics[width=0.48\textwidth]{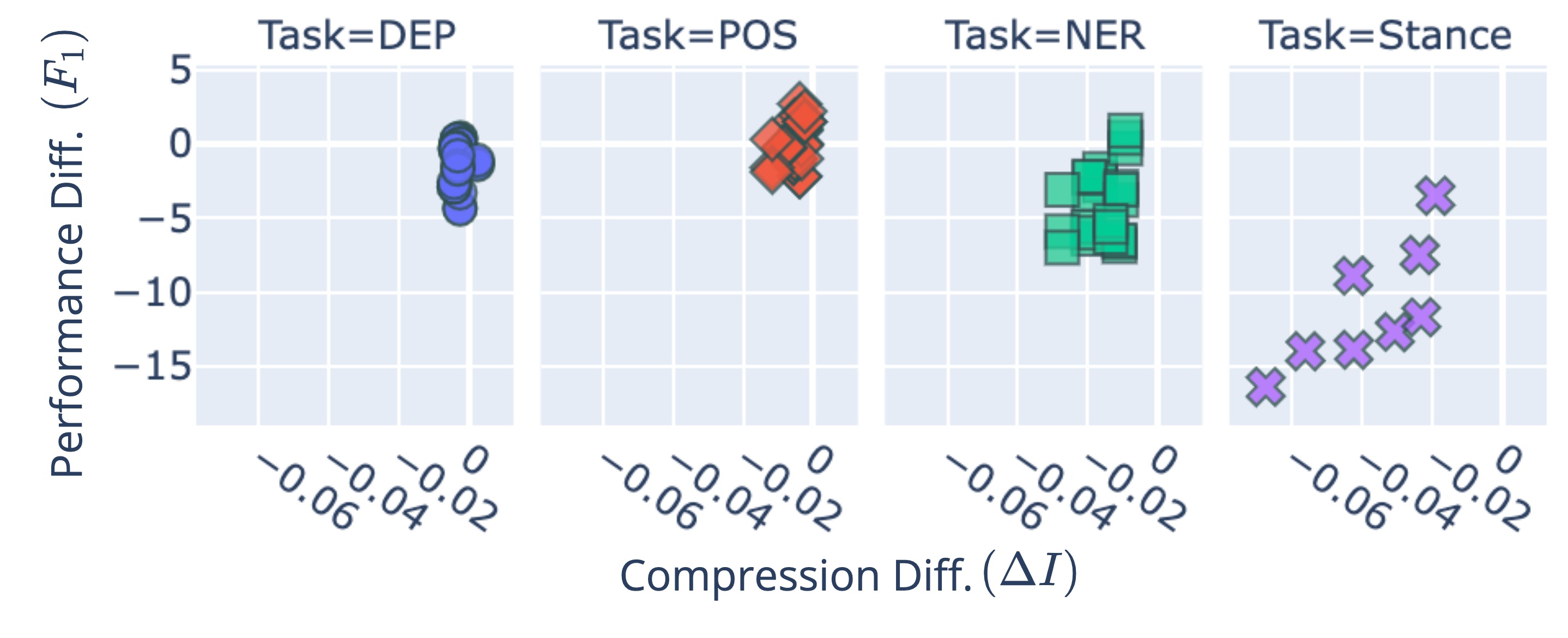}
  \caption{Comparision of the difference in $\Delta F_1$ and $\Delta I$ between Cross-Topic and In-Topic for all eight LMs on the four probing tasks.}
  \label{fig:exp1-diff}
\end{figure}

The first experiment shows that the generalization gap already exists after pre-training and varies regarding specific LMs and probing tasks. 
We analyze general (\autoref{tab:exp1} and \autoref{fig:exp1-diff}) and fine-grained (\autoref{tab:exp1-seen-unseen}) results and discuss them for the different evaluating setups, probing tasks, and LMs.
While firstly focusing on mid-size LMs usable for fine-tuning, we close how probing performance scales to large LMs in \autoref{sec:experiment-llm}.

\paragraph{Design}
We probe eight LMs on the probing tasks DEP, POS, NER, and Stance and verify them by observing significant performance drains using random initialized LMs (Appendix \autoref{sec:appendix-random-init}). 
For a holistic evaluation, we provide general results and group instances into two categories: \textit{seen} and \textit{unseen}.
We define \textit{seen} instances as already processed during training but in another context.
For example, the pronoun \textit{he} might appear in both training and test data, but in distinct sentences.
By evaluating the LMs on \textit{seen} instances, we gain insights into the influence of token-level lexical information versus context information from surrounding tokens.
In contrast, \textit{unseen} instances were not encountered during the training of a probe. 
They allow assessing whether LMs generalize to tokens that are similar to some extent (such as \textit{Berlin} and \textit{Washington}) but not seen during training.

\begin{table}[t]
\centering
    \resizebox{0.48\textwidth}{!}{%
  \setlength{\tabcolsep}{2pt}
\begin{tabular}{l@{\htc3}cc@{\htc3}cc@{\htc3}cc@{\htc3}cc@{\htc3}:ccc}
\toprule
 & \multicolumn{2}{c@{\htc3}}{\textbf{DEP}} & \multicolumn{2}{c@{\htc3}}{\textbf{POS}} & \multicolumn{2}{c@{\htc3}}{\textbf{NER}} & \multicolumn{2}{c@{\htc3}}{\textbf{Stance}}  & \multicolumn{3}{c}{\textit{Average}} \\\midrule
 & \textit{In} & \textit{Cross}&  \textit{In} & \textit{Cross}&  \textit{In} & \textit{Cross}&  \textit{In} & \textit{Cross}&  \textit{In} & \textit{Cross} & $\Delta$\\
\midrule
\ALBERT & \textbf{43.8} & \textbf{39.5} & \textbf{80.2} & \textbf{78.0} & \textbf{48.6} & 45.8 & 54.8 & \textbf{45.9} & \textit{\textbf{56.9}} & \textit{\textbf{52.3}}  & \textit{-4.6}\\
\BART & 36.5 & 36.9 & 75.4 & 74.1 & \textbf{48.7} & \textbf{45.3} & \textbf{60.8} & 44.4 & \textit{55.3} & \textit{50.2} & \textit{-5.1}\\
\BERT & 25.4 & 25.6 & 68.5 & 67.5 & 45.4 & 41.6 & 56.9 & 43.0 & \textit{49.0} & \textit{44.4}& \textit{-4.6}\\
\DEBERTA & 32.8 & 29.9 & 73.7 & 74.6 & \textbf{48.8} & 42.4 & \textbf{59.8} & \textbf{45.8} & \textit{53.4} & \textit{48.2} & \textit{-5.2}\\
\ROBERTA & 25.1 & 23.6 & 64.0 & 65.5 & \textbf{48.4} & 42.1 & 51.8 & 40.1 & \textit{47.3} & \textit{42.8} & \textit{-4.5}\\
\ELECTRA & 33.6 & 33.6 & 75.3 & 75.3 & 41.5 & 41.2 & 46.6 & 43.1 & \textit{49.3} & \textit{48.3}& \textit{\textbf{-1.0}}\\
\GPT & 25.2 & 23.9 & 63.5 & 61.9 & 45.5 & 38.6 & 51.1 & 38.4 & \textit{46.3} & \textit{40.7}& \textit{-5.6}\\
\GLOVE & 12.1 & 11.9 & 26.5 & 26.2 & 43.4 & 37.5 & 41.6 & 34.1 & \textit{30.9} & \textit{27.4}& \textit{-3.5}\\
\hdashline
\textit{Avg.} $\Delta$  & \multicolumn{2}{c@{\htc3}}{\textit{-1.2}}  & \multicolumn{2}{c@{\htc3}}{\textit{-0.5}} & \multicolumn{2}{c@{\htc3}}{\textit{-4.5}} & \multicolumn{2}{c@{\htc3}}{\textit{-11.0}}  & - & - &-\\
\bottomrule
\end{tabular}
}
  \caption{
In- and Cross-Topic probing results for eight LMs.
  We report the macro $F_1$ over three random seeds, the average difference between the two setups (last row), and their average per LM (last three columns). The best results within a gap of 1.0 are marked by columns.} 
\label{tab:exp1}
\end{table}

\begin{table}[t]
\centering
    \resizebox{0.48\textwidth}{!}{%
  \setlength{\tabcolsep}{2pt}
\begin{tabular}{ll@{\htc3}ccc@{\htc3}ccc@{\htc3}ccc}
\toprule
{} &  & \multicolumn{3}{c@{\htc3}}{\textbf{DEP}} & \multicolumn{3}{c@{\htc3}}{\textbf{POS}} & \multicolumn{3}{c@{\htc3}}{\textbf{NER}} \\ \midrule
{} &  & \textit{all} & $\Delta$ \textit{seen} & $\Delta$ \textit{unseen} & \textit{all} & $\Delta$ \textit{seen} & $\Delta$ \textit{unseen} & \textit{all} & $\Delta$ \textit{seen} & $\Delta$ \textit{unseen} \\ \midrule
 & \textit{Instance Ratio} & - & 85\% & 15\% & - & 86\% & 14\% & - & 65\% & 35\% \\\hdashline\noalign{\vskip 0.5ex}
 \multirow{5}{*}{\rotatebox[origin=c]{90}{In-Topic}}  & \ALBERT & 43.8 & +0.21 & -3.2 & 80.2 & +0.41 & -17.7 & 48.6 & +1.1 & -5.8 \\
 & \BART & 36.5 &  +0.13 & -3.0 & 75.4 & +0.20 & -16.5 & 48.7 & +1.3 & -7.0 \\
 & \BERT & 25.4 & -0.02 & -0.8 & 68.5 & +0.20 & -16.5 & 45.4 & +1.0 & -5.8 \\
 & \DEBERTA & 32.8 & +0.07 & -1.5 & 73.7 & +0.09 & -12.7 & 48.8 & +1.0 & -5.6 \\
 & \ROBERTA & 25.1 & -0.01 & -0.9 & 64.0 & -0.04 & -15.5 & 48.4 & +1.0 & -5.7  \\\hdashline\noalign{\vskip 0.5ex}
 & \textit{Average} & - & \textit{-0.08} & \textit{-1.9} & - & \textit{+0.17} &\textit{ -15.8} &-  & \textit{+1.1} & \textit{-6.0} \\\midrule
 & \textit{Instance Ratio} & - & 78\% & 22\% & - & 81\% & 19\% & - & 51\% & 49\% \\ \hdashline\noalign{\vskip 0.5ex}
 \multirow{5}{*}{\rotatebox[origin=c]{90}{Cross-Topic}}  & \ALBERT & 39.5 & +0.03 & -2.3 & 78.0 & +0.51 & -12.9 & 45.8 & +2.2 & -5.3 \\
 & \BART & 36.9 & +0.01 & -4.0 & 74.1 & +0.24 & -16.5 & 45.3 & +2.4 & -5.8 \\
 & \BERT & 25.6 & -0.09 & -0.7 & 67.5 & +0.20 & -14.0 & 41.6 & +1.9 & -5.1 \\
 & \DEBERTA & 29.9 & -0.07 & -1.3 & 74.6 & +0.14 & -11.7 & 42.4 & +2.0 & -5.2 \\
 & \ROBERTA & 23.6 &  -0.22 & -0.3 & 65.5 & +0.00 & -14.7 & 42.1 & +1.9 & -5.2  \\\hdashline\noalign{\vskip 0.5ex}
 & \textit{Average} & - & \textit{-0.08} & \textit{-1.7} & - & \textit{+0.22} &\textit{-14.0} &-  & \textit{+2.1} & \textit{-5.3} \\
\bottomrule
\end{tabular}
}
  \caption{Performance difference of \textit{seen} and \textit{unseen} instances compared to the full set (\textit{all}).
  We report for \ALBERT, \BART, \BERT, \DEBERTA, \& \ROBERTA, and include the ratio of \textit{seen} and \textit{unseen} instances.
  }
\label{tab:exp1-seen-unseen}
\end{table}

\paragraph{Results for Evaluation Setups}

Upon analyzing \autoref{tab:exp1}, we observe clear generalization gaps between In- and Cross-Topic evaluation for all tasks and LMs. 
As in \autoref{fig:exp1-diff}, the magnitude of this gap ($\Delta F_1$) correlates with the difference in compression ($\Delta I$). 
Interestingly, we find a stronger correlation between $F_1$ and $I$ for Cross-Topic ($\rho = 0.72$) as compared to In-Topic ($\rho = 0.69$). 
Thus, a higher performance level, like for In-Topic, leaves less room for compression improvements. 

Further, we examine the performance of \textit{seen} and \textit{unseen} instances in \autoref{tab:exp1-seen-unseen}.
It shows that \textit{seen} performs slightly better than \textit{all}, while \textit{unseen} ones underperform the complete set (\textit{all}) and \textit{seen} instances. 
Considering the average over LMs, there are fewer relative gains for \textit{seen} for In-Topic and more loss for \textit{unseen} instances (+1.2, -6.0 for NER) compared to Cross-Topic (+2.0, -5.3 for NER). 
This observation relates to the lower percentage of \textit{unseen} instances (i.e., made of topic-specific terms) for In- compared to Cross-Topic. 
We see \textit{unseen} instances on In-Topic are harder and cover rare vocabulary, and \textit{seen} instances on Cross-Topic are easier and made of general terms - which confirm our theoretical and semantic assumptions (\autoref{sec:topic-information}).

\paragraph{Results for Probing Tasks}

Considering \autoref{tab:exp1} and \autoref{fig:exp1-diff}, we note higher generalization gaps (\textit{Avg.} $\Delta$ of -4.5 and -11.0) for semantic tasks (NER and Stance) than for syntactic ones (DEP and POS) - \textit{Avg.} $\Delta$ of -1.2 and -0.5.
We verify this trend with results by observing a more pronounced gap for semantic NER classes (like ORG) than for syntactic ones (like ORDINAL) in the Appendix (\autoref{sec:appendix-ner}).

Next, we separately compare tasks for \textit{seen} and \textit{unseen} instances.
DEP shows the slightest performance difference compared to \textit{all}. 
We assume that the pairwise nature of the task leads to a larger shared vocabulary between \textit{unseen} and training instances - since a pair can be \textit{unseen}, but it may contain a frequent word like \textit{of}. 
In contrast, apparent differences between NER and POS are visible - with less performance drain on \textit{unseen} instances for NER than POS.
Therefore, we assume for NER a higher semantic overlap with training instances since they could include - as being an n-gram - words from the training vocabulary. 
In contrast, tokens of \textit{unseen} POS instances are always single words; thus, we assume a smaller semantic overlap with the training. 

\paragraph{Results for Encoding Models}

We now compare LMs amongst themselves. 
The four best-performing LMs of In-Topic differ up to 7.6 (\ALBERT~- \BERT), while for Cross-Topic, this difference narrows to 4.1 (\ALBERT~- \ELECTRA).
These results confirm the varying generalization gap between them and, again, that we can not transfer conclusions from one evaluation setup to another. 
For example, the probing performance of \BART~for In-Topic Stance is the best and the third best for Cross-Topic.

Generally, we do not see a clear correlation between better average performance and a smaller generalization gap. 
LMs like \DEBERTA~ perform better for In- and Cross-Topic but show a bigger gap (-5.1) compared to lower performing LMs like \ELECTRA~(-1.0), but there are also worse LMs with a bigger gap (\GPT, -5.6) or better ones with a smaller gap (\ALBERT, -4.6).
Overall, we see the generalization gap being more pronounced for better-performing LMs. 

Considering absolute performance, \ALBERT~and \BART~performs the best for both evaluation setups, while \ELECTRA~excels POS and DEP, and \DEBERTA~performs for NER and Stance.
In contrast, \BERT, \ROBERTA, \GPT, and \GLOVE underperform the others.
Thus, LMs with architectural regularization, such as layer-wise parameter sharing (\ALBERT), encoder-decoder layers (\BART), disentangled attention (\DEBERTA), or discriminator (\ELECTRA), tend to provide higher Cross-Topic performance.
Similarly, \ALBERT or \DEBERTA generally achieve more performance gains for \textit{seen} instances and fewer performance drops for \textit{unseen} ones than models without regularization such as \BERT~or \ROBERTA. 
We hypothesize that architectural and regularization aspects give LMs a more generalizable and robust encoding space.

\paragraph{Results for Larger Models}
\label{sec:experiment-llm}

\begin{table}[t]
\centering
    \resizebox{0.48\textwidth}{!}{%
  \setlength{\tabcolsep}{2pt}
\begin{tabular}{l@{\htc3}cc@{\htc3}cc@{\htc3}cc@{\htc3}cc@{\htc3}:ccc}
\toprule
 & \multicolumn{2}{c@{\htc3}}{\textbf{DEP}} & \multicolumn{2}{c@{\htc3}}{\textbf{POS}} & \multicolumn{2}{c@{\htc3}}{\textbf{NER}} & \multicolumn{2}{c@{\htc3}}{\textbf{Stance}}  & \multicolumn{3}{c}{\textit{Average}} \\\midrule
 & \textit{In} & \textit{Cross}&  \textit{In} & \textit{Cross}&  \textit{In} & \textit{Cross}&  \textit{In} & \textit{Cross}&  \textit{In} & \textit{Cross} & $\Delta$\\
\midrule
\ALBERT &43.8 & 39.5 & \textbf{80.2} & 78.0 & 48.6 & 45.8 & 54.8 & 45.9 & \textit{56.9} & \textit{52.3} & \textit{-4.6}\\
\BART & 36.5 & 36.9 & 75.4 & 74.1 & 48.7 & 45.3 & 60.8 & 44.4 & \textit{55.3} & \textit{50.2} & \textit{-5.1}\\\midrule
PYTHIA (12B) & 38.3 & 35.4 &  79.5 & 77.7 & 57.3 & 50.5 & 65.2 & 41.6 & \textit{60.1} & \textit{51.3} &  \textit{-8.8}\\
PYTHIA-DD (12B) & \textbf{45.3} & \textbf{45.4} & 79.8 & 79.2 & \textbf{64.5} & \textbf{55.8} & 66.1 & \textbf{50.4} &  \textit{\textbf{63.4}} & \textit{\textbf{57.9}} & \textit{-6.2}\\
\midrule
LLAMA-2 (13B) & \textbf{44.4} & 41.8 & \textbf{81.0} & \textbf{80.6} & 48.7 & 45.3 & \textbf{66.8} & 44.2 & \textit{60.2} & \textit{53.0} & \textit{-7.2}\\
LLAMA-2 Chat (13B) & \textbf{45.4} & 41.7 & \textbf{80.7} & \textbf{80.1} & 49.2 & 42.9 & \textbf{67.2} & 43.2 & \textit{60.6} & \textit{52.0} & \textit{-8.7}\\
\midrule
\bottomrule
\end{tabular}
}
  \caption{
Results (macro $F_1$) of the four probing tasks using the two overall best-performing LMs (\ALBERT~and \BART) in the In- and Cross-Topic setup based on the \textit{ArgMin} dataset (\autoref{tab:exp1}) and three large LMs.
  } 
\label{tab:llm-ukp-sent}
\end{table}

We compare in \autoref{tab:llm-ukp-sent} four open accessible large LMs with the two best performing models (\ALBERT~and \BART). 
In general, we see the performance scales with the higher number of parameters, but more noticeable for In- than Cross-Topic tasks. 
Therefore, the generalization gap of large LMs tend to be bigger than for LMs. 
Regarding the different large LMs, PYTHIA \citep{biderman2023pythia} and LLAMA-2 \citep{touvron2023llama} outperform the others on In-Topic tasks while performing on par with \ALBERT.
Further, we notice data deduplication during pre-training (PYTHIA-DD) results in the best performing model and actively reduces the generalization gap from $8.8$ to $6.2$.
In addition, instruction fine-tuning does not heavily affect the performance but tends to increase the generalization gap from $7.2$ (LLAMA-2) to $8.7$ (LLAMA-2 Chat).

\begin{figure*}[]
  \centering
  \includegraphics[width=1\textwidth]{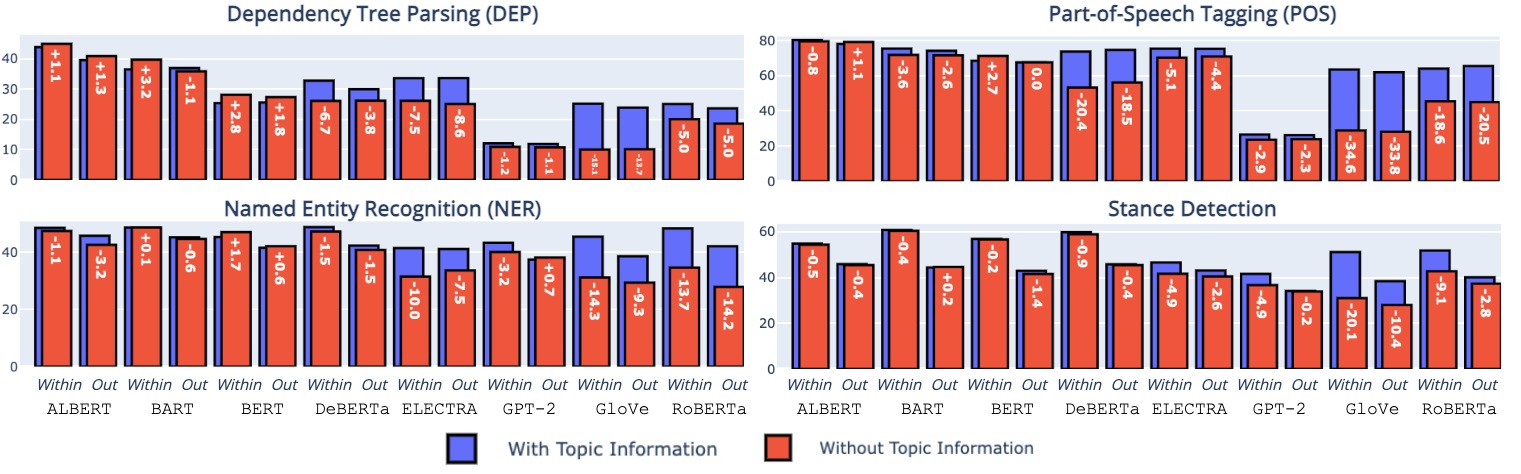}
  \caption{Comparison of the probing results with (blue bars) or without (red bars) topic information. The white text indicates the difference between these two scenarios ($\Delta F_1^{\setminus T}$).}
  \label{fig:diff}
\end{figure*}

\section{The Dependence on Topic-Specific Vocabulary}
\label{sec:experiment-2}


To this point, we saw that the generalization gap varies between different LMs and probing tasks. 
Since topic-specific vocabulary crucially affects generalization gaps, we analyze the varying dependence on the topic-specific vocabulary of LMs using \textit{Amnesic Probing} \citep{elazar-etal-2021-amnesic}.
We observe apparent differences among LMs and assume their embedding space clearly differs beyond single evaluation metrics.
Therefore, we emphasize considering various LMs when using \textit{Amnesic Probing}.
Additional insights of comparing \textit{seen} and \textit{unseen} instance and distinct NER classes are provided in the Appendix (\autoref{sec:appendix-amnesic-seen-unseen}, \autoref{sec:appendix-ner-without}).

\paragraph{Design}
To measure how LMs depend on topic-specific vocabulary, we employ \textit{Amnesic Probing} \citep{elazar-etal-2021-amnesic} to remove the latently encoded topic-specificity $z_i$ from the embeddings $h_i$ of a token $w_i$.  
More precisely, we compare how the performance of a probing task (like NER) changes when we remove $z_i$.
A more negative effect indicates a higher dependence on topic-specific vocabulary, while this property is a hurdle when performance improves. 
We first train a linear model on token-level topic-specificity $r$ (\autoref{para:token-level-topic-information}).
To shape it as a classification task, we categorize $r$ into three classes (\textit{low}, \textit{medium}, \textit{high}). \footnote{Please find examples in the Appendix \autoref{appendix:token-level-examples}.}
Next, we find a projection matrix $P$ that projects all embeddings $h_i$ - gathered as $H$ - using the learned weights $W_l$ of $l$ to the null space as $W_l P H = 0$.
Using $P$ we update $h_i$ by neutralizing topic-specificity from the input as $h_i^{'} = P h_i$ before training the probe. 
Following \citep{elazar-etal-2021-amnesic}, we verified our results by measuring less effect of removing random information from $h_i$ (see Appendix \autoref{sec:appendix-random-information}).

\paragraph{Results}

Considering \autoref{fig:diff}, we see \ALBERT, \BART, and \BERT~depend less on topic-specific vocabulary.
Their diverse pre-training (token- and sentence-objectives or sentence denoising) results in a more robust embedding space. 
Surprisingly, they show positive effects (3.2 for DEP for \BART) when removing topic-specificity.
This could remove potentially disturbing parts of the embedding space. 
Similarly, \GPT~ is less affected by the removal - we assume this is due to its generally lower performance level.
Therefore, it has less room for performance drain, and capturing topic-specificity is less powerful.

Comparing In- and Cross-Topic setups shows a narrowing generalization gap for more affected models (like \ROBERTA~and \GLOVE~on NER or NER).
Simultaneously, less affected LMs either maintain the gap or enlarge it slightly - like \BART~on DEP, NER, or NER. 
Further, \DEBERTA, \ROBERTA, \ELECTRA, and \GLOVE~rely more on topic-specific vocabulary since they show significant performance loss (up to 34.6 for \GLOVE~on POS) when removing this information.
Specifically, \GLOVE~as a static language model, and \ROBERTA~is affected the highest for all tasks. 
\ELECTRA~shows similar behavior but is less pronounced for POS. 
Thus, its reconstruction pre-training objective provides a more robust embedding space than purely MLM (\DEBERTA~or \ROBERTA).
Comparing \DEBERTA~and \ROBERTA, \DEBERTA~is less affected by the removal of semantic tasks (NER and NER).
We hypothesize that distinguishing between token content and token position via disentangled attention makes \DEBERTA~more robust for the semantic than for syntactic tasks (DEP and POS).

\section{The Evolution of the Generalization Gap during Fine-Tuning}
\label{sec:experiment-3}

Finally, we re-evaluate fine-tuned LMs using our proposed probing setups and show that fine-tuning leads to a drain in probing performance.
We use these results to retrace apparent differences between evaluation setups and the varying generalization gap between LMs.
This is relevant for a broader understanding of how fine-tuning affects LMs \citep{mosbach-etal-2020-interplay-fine, DBLP:conf/iclr/KumarRJ0L22}, and what they learn during fine-tuning \citep{DBLP:conf/coling/MerendiDV22,DBLP:conf/eacl/RavichanderBH21}. 

\paragraph{Design}
We fine-tune the LMs on an argumentative \emph{stance detection} task and re-evaluate them on DEP, POS, and NER probing tasks.
To be consistent with our probing setup, we used the same folds for fine-tuning.
Further details are in the Appendix (\autoref{sec:fine-tuning-details}).
We compare these results with the probing performance of their pre-trained counterparts (\autoref{sec:experiment-1} and \autoref{sec:experiment-2}) and correlate this change with the generalization gap observed on the downstream task.
We limit our analysis to \ALBERT,~\BERT,~\BART,~\DEBERTA, and \ROBERTA.

\begin{table}[]
\centering
    \resizebox{0.48\textwidth}{!}{%
  \setlength{\tabcolsep}{3pt}
\begin{tabular}{llc@{\htc3}cccc@{\htc3}ccc}
\toprule
&   & \textit{\textbf{Stance}} &\textbf{DEP} & \textbf{POS} & \textbf{NER}  & \textbf{\textit{Avg.}}  & \textbf{DEP} & \textbf{POS} & \textbf{NER} \\\midrule
  & & $F_1$ \textit{fine-tuned}  &\multicolumn{4}{c@{\htc3}}{$\Delta F_1$ \textit{probing}} & \multicolumn{3}{c@{\htc3}}{$\Delta F_1^{\setminus T}$} \\
\midrule
\multirow{5}{*}{\rotatebox[origin=c]{90}{In-Topic}} & \ALBERT & 55.4  \small{+0.6} & -27.3 & -40.2 & -25.0&  -30.8 & -0.6 & -3.0 & -0.1   \\
 & \BART & 69.8  \small{+9.0} & -17.3 & -32.2 & -4.0 & -17.8 & -0.8 & -4.0 & +0.3   \\
 & \BERT & 67.2 \small{+10.3} & -7.5 & -24.8 & +1.0 & -10.4 & +0.4 & +0.7 & +1.1   \\
 & \DEBERTA & \textbf{70.1  \small{+10.3}} & -13.2 & -25.3 & -8.8 & -15.8 & -0.8 & -3.8 & -0.4 \\
 & \ROBERTA & 68.9  \small{+17.1} & -19.7 & -48.6 & -29.7 & -27.2 & -0.8 & -3.0 & -0.7 \\\hdashline\noalign{\vskip 0.5ex}
  & \textit{Avg.} & 66.3 \small{+9.5} & -16.6 & -32.6  & -12.1 & -20.4 & -0.5 & -2.6 & +0.1  \\
\midrule
 \multirow{5}{*}{\rotatebox[origin=c]{90}{Cross-Topic}} & \ALBERT& 51.4  \small{+5.5} & -14.4 & -20.3 & -12.6 &  -15.8 & +1.6	& -1.3	& +2.1  \\
 & \BART& \textbf{61.9}  \textbf{\small{+17.5}} & -16.5 & -33.9 & -5.4 & -18.6 & -1.0 & -3.5 & -1.6  \\
 & \BERT& 56.6 \small{+13.6} & -5.7 & -19.5 & +0.6 & -8.2 & +0.7 & +0.6 & +1.2 \\
 & \DEBERTA& 55.9  \small{+10.1} & -13.4 & -33.4 & -11.8 & -19.5 & -1.2 & -8.6 & +1.6 \\
 & \ROBERTA & 55.5  \small{+15.4} & -16.6 & -48.3 & -23.1 & -23.5 & -1.9 & -4.8 & -0.3 \\\hdashline\noalign{\vskip 0.5ex}
  & \textit{Avg.} & 56.3 \small{+12.6} & -13.0 & -29.3 & -9.1 & -17.1 & -0.4 & -3.5 & +0.6 \\
\bottomrule
\end{tabular}
}
  \caption{
  Results of evaluating our probing setup on fine-tuned LMs on NER.
The first column shows these fine-tuned results and the gained improvement compared to probing for NER on pre-trained LMs (\autoref{tab:exp1}).
  Next, we show performance differences between pre-trained and fine-tuned LMs ($\Delta F_1$ \textit{probing}) and how removing topic-specificity affects the fine-tuned LMs ($\Delta F_1^{\setminus T}$). 
  }
\label{tab:exp3-overall}
\end{table}

\paragraph{Results}

\autoref{tab:exp3-overall} shows that fine-tuning clearly boost the performance on NER compared to the probing performance (\autoref{sec:experiment-1}) but leads to a clear performance drop ($\Delta F_1$) for both evaluation setups and the probing tasks.
Cross-Topic achieved more gains on average (+12.6) and fewer drains (-17.1) on the three linguistic properties than In-Topic (+9.5, -20.4).
On average, we assume that In-Topic fine-tuning affects the encoding space of LMs more heavily than Cross-Topic. 
Regarding the different probing tasks, the performance drain is more pronounced for syntactic tasks (DEP and POS) than semantic tasks (NER).
This hints that LMs acquire competencies of a semantic nature - which holds for \emph{stance detection}. 
Similarly, removing topic-specificity influences fine-tuned LMs the least for NER.
At the same time, this removal is more pronounced for Cross-Topic.
This confirms the assumption that the Cross-Topic setup has smaller effects on LMs internals since we saw big impacts of this removal  (\autoref{sec:experiment-2}).

Considering the single LMs, we see apparent differences. 
For example, \ALBERT, with its shared architecture and priorly best-performing LM, experiences big probing performance drains and the smallest fine-tuning gains (+0.6, +5.5).
In contrast, we note effective fine-tuning of \BERT with +10.3 for In- and +13.6 for Cross-Topic, and that it lost the least probing performance.
Comparing \ROBERTA~and \DEBERTA~reveals again the effectiveness of architectural regularization of \DEBERTA.
\ROBERTA~ shows the most gains when fine-tuning on NER and almost catching up with \DEBERTA.
However, it experiences a more clear performance drain (-27.2, -23.5) regarding the probing tasks for In- and Cross-Topic compared to \DEBERTA~(-15.8, -19.5).
Next, we focus on \BART~ and its superior Cross-Topic performance on NER.
It seems already well-equipped for this downstream task due to its high In-Topic probing performance on NER.
Therefore, it can learn the task more robustly during fine-tuning.  


\section{Related Work}
\label{sec:related_work}

The rise of LMs \citep{devlin-etal-2019-bert, Liu2019RoBERTaAR, radford2019language, he2021deberta} enabled big success on a wide range of tasks \citep{wang-etal-2018-glue, NEURIPS2019_4496bf24}. Nevertheless, they still fall behind on more realistic Cross-Topic, like generalizing towards unseen topics \citep{stab-etal-2018-cross, gulrajani2020search, allaway-mckeown-2020-zero}. 
One primary reason is that LMs often rely on unwanted spurious correlations.
Despite LMs seeing such vocabulary during pre-training, they failed to consider test vocabulary in the required fine-grained way \citep{thorn-jakobsen-etal-2021-spurious, reuver-etal-2021-stance}. 
Further, \newcite{Kumar2022FineTuningCD} found linear models can outperform fine-tuning LMs when considering out-of-distribution data.
Thus, a broader understanding of LMs in challenging evaluation setups is crucial.

Probing \citep{belinkov-etal-2017-neural, conneau-etal-2018-cram, peters-etal-2018-dissecting} helps to analyze inners of LMs.
This includes to examine how linguistic \citep{tenney-etal-2019-bert,Tenney2019WhatDY}, numeric \citep{Wallace2019Numbers}, reasoning \citep{10.1162/tacl_a_00342}, or discourse \citep{Koto2021DiscoursePO} properties are encoded.
Other works focus on specific properties used for other tasks \citep{elazar-etal-2021-amnesic,lasri-etal-2022-probing}, or fine-tuning dynamics \citep{merchant-etal-2020-happens,zhou-srikumar-2022-closer,Kumar2022FineTuningCD}. 
However, these works target the commonly used \textit{In-Topic} setup and less work considering Cross-Topic setups.
\newcite{aghazadeh-etal-2022-metaphors} analyzed metaphors across domains and language, or \newcite{zhu2022ood} cross-distribution probing for visual tasks. 
They found that models generalize to some extent across distribution shifts in probing-based evaluation.  
Nevertheless, these works focus on specialized tasks and consider the generalizations across distributions in isolation.
In contrast, we propose with our experiments a more holistic probing-based evaluation of LMs, covering different generalization aspects after pre-training and fine-tuning. 

\section{Conclusion}

\paragraph{Discussion}

We analyzed and compared In- and Cross-Topic evaluation setups and found generalization gaps significantly differing regarding specific LMs and probing tasks.\footnote{We verified our results using a second dataset from the social media domain \citep{conforti-etal-2020-will} - details in the Appendix \autoref{sec:appendix-generalization}.}
Further, we make various crucial observations contributing to a better understanding of the generalizability of LMs:
\textbf{(1)} diverse pre-training objectives and architectural regularization tend to positively affect the robustness of LMs and their embedding space, such as depending less on topic-specific vocabulary; \textbf{(2)} probing performance falls short for rare vocabulary, underscoring the need to explore token-level properties; \textbf{(3)} probing performance, but also generalization gaps, tend to scale for larger LMs, while deduplication of pre-training data improves their robustness and narrows these gaps; and \textbf{(4)} In-Topic fine-tuning tend to vanish linguistic properties more prominently than for the Cross-Topic setup. 

To conclude, we highlight the practical utility of probing to analyze and compare the capacities of various LMs from a different perspective - considering different generalization scenarios. 
Thereby, our work points out the importance of probing as a universally applicable method, regardless of size or being static or contextualized, to complement existing work on analyzing language models \citep{wang-etal-2018-glue,DBLP:journals/corr/abs-2211-09110}.

\paragraph{Outlook}
With our findings in mind, we regularly see probing LMs and large LMs and consider forthcoming learning paradigms as indispensable for a holistic evaluation of their verity and multiplicity. 
Therefore, we will continue to analyze language models, including a broader set of tasks and focusing on general and rare vocabulary to increase our understanding of how, why, and where they differ.

\section*{Acknowledgements}
We thank Irina Bigoulaeva, Tim Baumgärtner, Tilman Beck, and the anonymous reviewers for their valuable feedback.
This work has been funded by the Hasler Foundation Grant No. 21024.

\section*{Ethical Considerations and Limitations}

\paragraph{Automatic Annotations for Linguistic Properties}
Our experiments require all instances origin in the same datasets with topic annotations.
Thanks to this condition, we align all our experiments, like probing LMs, with the same data as they got pre-trained.
Therefore, we minimize other influences like semantic shifts of other datasets.
However, there are no corresponding annotations for linguistic properties, which forces us to rely on automatically gathered annotations. 
This work addresses this issue by transparently stating the libraries and models we used to derive these annotations and providing the source code and the extracted labels in our repository.
We compared our results (\autoref{sec:appendix-label-verification}) with previous work \citep{tenney-etal-2019-bert,Tenney2019WhatDY,hewitt-liang-2019-designing} and found our results well aligned. 
Further, we verify the probing task results on the different LMs with randomly initialized counter-parts (\autoref{sec:appendix-random-init}) and confirm our findings with a second dataset (\autoref{sec:appendix-generalization}). 

\paragraph{Definition of Topic-Specific Vocabulary}
This work considers a topic as a semantic grouping provided by a given dataset.
As previously mentioned, this focus on the context of one dataset allows in-depth and controlled analysis, like examining the change of LMs during fine-tuning. 
On the other hand, we need to re-evaluate other datasets since the semantic space and granularity of the topic are different in almost every other dataset. 
Nevertheless, results in the Appendix (\autoref{sec:appendix-generalization}) let us assume that our findings correlate with other datasets and domains.
Further, we consider only token-level specific vocabulary, as done previously in literature \citep{kawintiranon-singh-2021-knowledge}.
We think that considering n-grams could give a better approximation of topic-specific terms. 
Still, we do not consider them because \textit{Amnesic Probing} \citep{elazar-etal-2021-amnesic} require token-level properties to apply resulting intervention on token-level tasks like POS.

\paragraph{Impact of LMs Design choices}
This work analyzes LMs regarding different properties like pre-training objectives or architectural regularization. 
However, we do not claim the completeness of these aspects nor a clear causal relationship.
Making such a final causal statement would require significant computational resources to pre-train models to verify single properties with full certainty. 
Instead, we use same-sized model variations, evaluate all probes on three folds and three random seeds to account for data variability and random processes, and verify our results on a second dataset. 
Nevertheless, we use them to correlate results on aggregated properties (like having diverse pre-training objectives or not) and not on single aspects, like the usefulness of the \textit{Sentence-Order} objective.

\bibliography{anthology,custom}
\bibliographystyle{acl_natbib}

\clearpage
\appendix

\section{Additional Details of the Experiments}

\subsection{Probing Tasks}\label{subsec:appendix:tasks}
\autoref{tab:tasks} shows examples and additional details of the different probing tasks.

\begin{table*}[]
\centering
    \setlength{\tabcolsep}{3pt}
    \resizebox{1\textwidth}{!}{%
    \begin{tabular}{lccccc}
    \toprule
     \bf Task & \bf Example & \bf Label & \bf \# Instances & \bf \# Labels \\
    \midrule
   \texttt{DEP} & I think there is a lot \underline{we} can \underline{learn} from Colorado and Washington State.  & \textit{nsubj} & 40,000   & 41 \\
    \texttt{POS} & I think there is a lot \underline{we} can learn from Colorado and \underline{Washington State}.  & \textit{PRON}  & 40,000 & 17  \\
      \texttt{NER} & I think there is a lot we can learn from Colorado and \underline{Washington State}.  &  \textit{PERS} & 25,892 & 17\\
  \texttt{Stance} & \underline{I think there is a lot we can learn from Colorado and Washington State.} & \textit{PRO}& 25,492 & 3  \\ 
    \bottomrule
    \end{tabular}
    }
    \caption{Overview and examples of the different probing tasks.}
    \label{tab:tasks}
\end{table*}

\subsection{Fold Composition}\label{subsec:appendix:folds}

\begin{figure}[]
  \centering
  \includegraphics[width=0.48\textwidth]{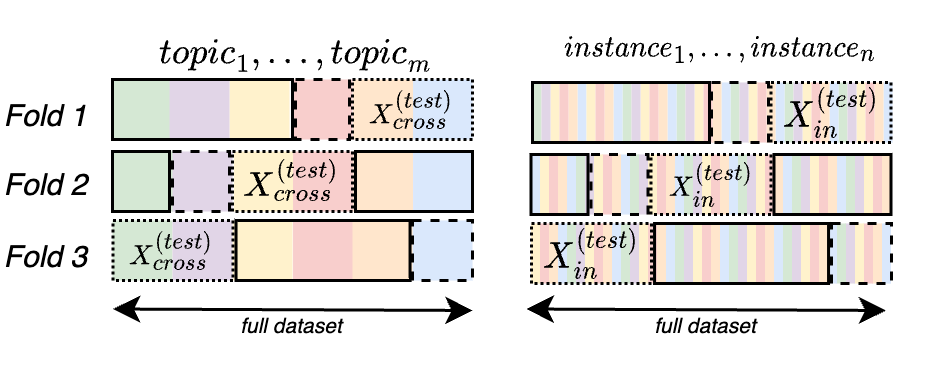}
  \caption{Overview of the In- and Cross-Topic setup using three folds. 
  The colour indicates a topic; solid lines train-, dotted lines dev-, and dashed lines test-splits.}
  \label{fig:folds}
\end{figure}

We rely on a three-folded evaluation for In- and Cross-Topic for a generalized performance measure. 
These folds cover every instance exactly once in a test split. 
In addition, we require that In- and Cross-Topic train/dev/test splits have the same number of instances for a fair comparison, as visualized in \autoref{fig:folds}.
For Cross-Topic, we make sure that every topic $\{t_1, ..., t_m\}$ is covered precisely once by one of the three test splits $X_{cross}^{(test)}$.
To compose $X_{cross}^{(train)}$ and $X_{cross}^{(dev)}$, we randomly distribute the remaining topics for every fold. 
For In-Topic, we randomly\footnote{We expect that all folds cover all topics given the small number of topics (8) and the big number of instances.} form subsequent test splits $X_{in}^{(test)}$ for every fold from all instances $\{x_1, ..., x_m\}$.
$X_{in}^{(train)}$ and $X_{in}^{(dev)}$ are then randomly composed for every fold using the remaining instance set following the dimension of $X_{cross}^{(train)}$ and $X_{cross}^{(dev)}$.

\subsection{Training Setup}
For all our experiments, we use NVIDIA RTX A6000 GPUs, python (3.8.10), transformers (4.9.12), and PyTorch (1.11.0).

\subsection{Probing Hyperparameters}
Further, we use for the training of the probes the following fixed hyperparameters: 20 epochs, where we find the best one using dev instances; AdamW \citep{adamW2019} as optimizer; a batch size of 64; a learning rate of 0.0005; a dropout rate of 0.2; a warmup rate of 10\% of the steps; random seeds: $[0,1,2]$

In addition, we use the following tags from the huggingface model hub:

\begin{itemize}
    \item \href{https://huggingface.co/albert-base-v2}{\texttt{albert-base-v2}}
    \item \href{https://huggingface.co/bert-base-uncased}{\texttt{bert-base-uncased}}
    \item \href{https://huggingface.co/facebook/bart-base}{\texttt{facebook/bart-base}}
    \item \href{https://huggingface.co/microsoft/deberta-base}{\texttt{microsoft/deberta-base}}
    \item \href{https://huggingface.co/roberta-base}{\texttt{roberta-base}}
    \item \href{https://huggingface.co/google/electra-base-discriminator}{\texttt{google/electra-base\-discriminator}}
    \item \href{https://huggingface.co/gpt2}{\texttt{gpt2}}
    \item \href{https://huggingface.co/EleutherAI/pythia-12b}{\texttt{EleutherAI/pythia-12b}}
    \item \href{https://huggingface.co/EleutherAI/pythia-12b-deduped}{\texttt{EleutherAI/pythia-12b-deduped}}
    \item \href{https://huggingface.co/meta-llama/Llama-2-13b-hf}{\texttt{meta-llama/Llama-2-13b-hf}}
    \item \href{https://huggingface.co/meta-llama/Llama-2-13b-chat-hf}{\texttt{meta-llama/Llama-2-13b-chat-hf}}
    \item \href{https://huggingface.co/google/t5-xxl-lm-adapt}{\texttt{google/t5-xxl-lm-adapt}}
    \item \href{https://huggingface.co/allenai/tk-instruct-11b-def}{\texttt{allenai/tk-instruct-11b-def}}

\end{itemize}

\subsection{Fine-Tuning Hyperparameters}
\label{sec:fine-tuning-details}

To fine-tune on \textit{stance detection}, we use the following setup:
5 epochs, where we find the best one using dev instances; AdamW \citep{adamW2019} as optimizer; a batch size of 16; a learning rate of 0.00002; a warmup rate of 10\% of the steps; random seeds: $[0,1,2]$.

\subsection{Token-Level Examples for Topic Relevance}
\label{appendix:token-level-examples}

In \autoref{sec:experiment-2}, we use the binned topic-specificity (\autoref{sec:experiment-2}) for each token.
We show in \autoref{tab:topic-information-examples} examples for three bins \textit{low}, \textit{medium}, and \textit{high}.
The first bin (\textit{low}) is made of tokens, which barely occur in the dataset. 
The second one (\textit{medium}) consists of tokens which are part of most topics.
Finally, the last bin (\textit{high}) includes tokens with a high topic relevance for ones like \textit{Cloning} or \textit{Minimum Wage}.

\begin{table}[]
\centering
    \resizebox{0.5\textwidth}{!}{%
  \setlength{\tabcolsep}{3pt}
\begin{tabular}{ccc}
\toprule
  \textit{\textbf{low}} & \textit{\textbf{medium}} &  \textit{\textbf{high}} \\ \midrule
fianc, joking, validate,& as, on, take, & cloning, uniform, wage, \\
latitude, poignantly, informative & some, like, how, & marijuana, minimum, gun, \\
ameliorate, bonding, mentors& so, one, these, & cloned, wear, clone,\\ 
brigade, emancipation, deriving,& instead, while, ago & nuclear, energy, penalty, \\
ignatius, 505, nominations, & where, came, still, many,  & uranium, legalization, cannabis, \\
electorate, SWPS, 731& come, engage, seems & execution, wast, employment \\
\bottomrule
\end{tabular}
}
  \caption{Examples of tokens with a \textit{low}, \textit{medium,} or \textit{high} token relevance following \autoref{sec:experiment-1}.
  }
\label{tab:topic-information-examples}
\end{table}

\section{Further Results}

\subsection{Generalization Across Datasets}
\label{sec:appendix-generalization}

\begin{figure*}[]
  \centering
  \includegraphics[width=1.0\textwidth]{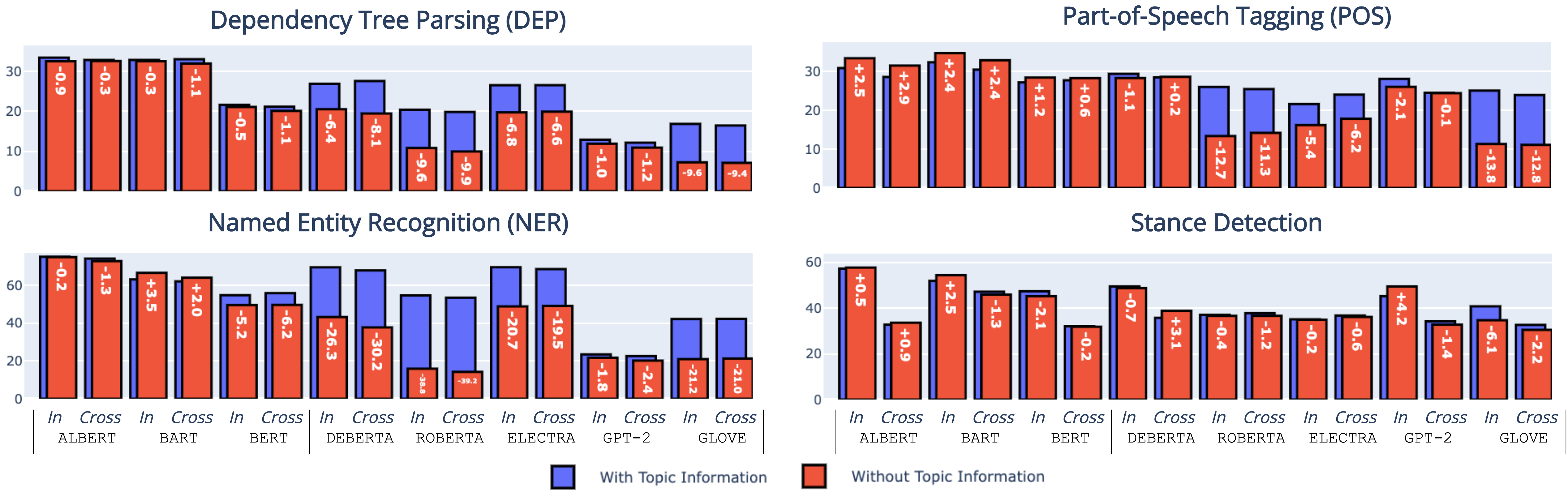}
  \caption{Comparison of the probing results with (blue bars) or without (red bars) topic-specificity for the \textit{will-they-wont-they} dataset \citep{conforti-etal-2020-will}. The white text indicates the difference between these two scenarios.}
  \label{fig:amnesic-overview-wtwt}
\end{figure*}

With \autoref{tab:wtwt}, and \autoref{fig:amnesic-overview-wtwt} we verify the results of \autoref{sec:experiment-1}, \autoref{sec:experiment-2}, and \autoref{sec:experiment-llm} using another \emph{stance detecion} dataset. 
Namely, we use the \textit{wtwt} (\textit{will-they-wont-they}) \citep{conforti-etal-2020-will} dataset which covers 51.284 tweets annotated either \textit{support}, \textit{refute}, \textit{comment}, or \textit{unrelated} towards five financial topics.   
The overall performance comparison between In- and Cross-Topic shows the same trend as we already saw in \autoref{sec:experiment-1}, but on a lower level. 
We assume this is mainly due to this dataset's more specific domain (twitter) compared to \textit{UKP ArgMin}.
Focusing on the influence of topic-specific vocabulary verifies the previously presented results (\autoref{sec:experiment-2}) again. 
LMs pre-trained with purely token-based objectives highly depend on topic-specific vocabulary. 
 
\begin{table}[]
\centering
    \resizebox{0.48\textwidth}{!}{%
  \setlength{\tabcolsep}{2pt}
\begin{tabular}{l@{\htc3}cc@{\htc3}cc@{\htc3}cc@{\htc3}cc@{\htc3}:ccc}
\toprule
 & \multicolumn{2}{c@{\htc3}}{\textbf{DEP}} & \multicolumn{2}{c@{\htc3}}{\textbf{POS}} & \multicolumn{2}{c@{\htc3}}{\textbf{NER}} & \multicolumn{2}{c@{\htc3}}{\textbf{NER}}  & \multicolumn{3}{c}{\textit{Average}} \\\midrule
 & \textit{In} & \textit{Cross}&  \textit{In} & \textit{Cross}&  \textit{In} & \textit{Cross}&  \textit{In} & \textit{Cross}&  \textit{In} & \textit{Cross} & $\Delta$\\
\midrule
\ALBERT & \textbf{33.5} & \textbf{32.9} & \textbf{75.1} & \textbf{74.2} & 30.9 & 28.6 & \textbf{57.3} & 32.8 & \textbf{\textit{49.1}} & \textit{42.1} & \textit{-7.0} \\
\BART & \textbf{32.9} & \textbf{33.1} & 63.2 & 62.1 & \textbf{32.4} & \textbf{30.5} & 51.9 & \textbf{47.2} &\textit{ 45.1} & \textbf{\textit{43.2}} &  \textit{-1.9}\\
\BERT & 21.6 & 21.2 & 54.8 & 55.9 & 27.2 & 27.8 & 47.4 & 32.1 & \textit{37.8} & \textit{34.2} &  \textit{-3.6}\\
\DEBERTA & 26.9 & 27.6 & 69.6 & 67.9 & 29.4 & 28.5 & 49.5 & 35.7 & \textit{43.9} & \textit{40.0} &  \textit{-3.9}\\
\ROBERTA & 20.4 & 19.9 & 54.7 & 53.5 & 26.1 & 25.5 & 37.0 & 37.8 & \textit{35.6} & \textit{34.2} &  \textit{-1.4}\\
\ELECTRA & 26.6 & 26.6 & 69.6 & 68.6 & 21.7 & 24.1 & 35.1 & 36.7 & \textit{38.2} & \textit{39.0} & \textbf{\textit{+0.8}} \\
\GPT2 & 16.9 & 16.5 & 42.2 & 42.2 & 25.1 & 24.0 & 40.8 & 32.6 &\textit{ 31.2} & \textit{28.8} &  \textit{-2.4}\\
\GLOVE & 12.9 & 12.2 & 23.5 & 22.6 & 28.1 & 24.6 & 45.2 & 34.2 &\textit{ 27.4} & \textit{23.4} &  \textit{-4.0}\\
\hdashline
\textit{Avg.} $\Delta$  & \multicolumn{2}{c}{\textit{-0.3}}  & \multicolumn{2}{c}{\textit{-0.7}} & \multicolumn{2}{c}{\textit{-0.9}} & \multicolumn{2}{c}{\textit{-9.5}}  & - & - &-\\
\bottomrule
\end{tabular}
}
  \caption{
Results of the four probing tasks using eight LMs in the In- and Cross-Topic setup.
  We report the mean $F_1$ (macro averaged) over three random seeds, the average difference between the two evaluation setups per task (last row), and their average per LM (last two columns). Best-performing results within a margin of 1pp are marked for every task and setup.} 
\label{tab:wtwt}
\end{table}

\subsection{Comparison of Probing Tasks against Random Initialized LMs}
\label{sec:appendix-random-init}

We show in \autoref{tab:random-in} and \autoref{tab:random-cross} the results of running the three linguistic probes on the seven contextualized LMs in their random initialized version. 
For In- and Cross-Topic, there is a clear performance drop of having random initialized models. 

\begin{table}[ht]
\centering
    \resizebox{0.48\textwidth}{!}{%
  \setlength{\tabcolsep}{3pt}
\begin{tabular}{l@{\htc3}cc@{\htc3}cc@{\htc3}cc}
\toprule
 & \multicolumn{2}{c@{\htc3}}{\textbf{DEP}} & \multicolumn{2}{c@{\htc3}}{\textbf{POS}} & \multicolumn{2}{c@{\htc3}}{\textbf{NER}} \\\midrule
 & \textit{Random} & $\Delta$ &  \textit{Random} &  $\Delta$ &  \textit{Random} & $\Delta$ \\
\midrule
\ALBERT              &        1.4 &      -42.4 &        6.8 &      -41.8 &        3.4 &      -76.8 \\
\BART                &        1.4 &      -35.1 &        5.0 &      -43.7 &        2.7 &      -72.7 \\
\BERT                &        2.7 &      -22.7 &        9.4 &      -36.0 &        4.6 &      -63.9 \\
\DEBERTA             &        7.0 &      -25.8 &       16.3 &      -32.5 &       16.1 &      -57.6 \\
\ROBERTA             &        2.2 &      -22.9 &       11.0 &      -37.4 &        4.7 &      -59.3 \\
\ELECTRA             &        1.7 &      -31.9 &        8.4 &      -33.1 &        3.8 &      -71.5 \\
\GPT                &        5.8 &      -19.4 &       12.3 &      -33.2 &       12.5 &      -51.0 \\
\bottomrule
\end{tabular}
}
  \caption{
Results of evaluating DEP, POS, and NER using the seven contextual LMs (random initialized) for In-Topic and the difference to their pre-trained counterparts in \autoref{tab:exp1}.} 
\label{tab:random-in}
\end{table}

\begin{table}[]
\centering
    \resizebox{0.48\textwidth}{!}{%
  \setlength{\tabcolsep}{3pt}
\begin{tabular}{l@{\htc3}cc@{\htc3}cc@{\htc3}cc}
\toprule
 & \multicolumn{2}{c@{\htc3}}{\textbf{DEP}} & \multicolumn{2}{c@{\htc3}}{\textbf{POS}} & \multicolumn{2}{c@{\htc3}}{\textbf{NER}} \\\midrule
 & \textit{Random} & $\Delta$ &  \textit{Random} &  $\Delta$ &  \textit{Random} & $\Delta$ \\
\midrule
\ALBERT              &        1.4 &      -38.1 &        6.2 &      -39.6 &        3.4 &      -74.6 \\
\BART                &        1.5 &      -35.4 &        5.0 &      -40.3 &        2.9 &      -71.2 \\
\BERT                &        2.1 &      -23.5 &        9.6 &      -32.0 &        4.5 &      -63.0 \\
\DEBERTA             &        6.8 &      -23.1 &       14.0 &      -28.4 &       17.2 &      -57.4 \\
\ROBERTA             &        2.6 &      -21.0 &       10.0 &      -32.1 &        5.2 &      -60.3 \\
\ELECTRA             &        3.0 &      -30.6 &        9.8 &      -31.4 &        4.1 &      -71.2 \\
\GPT                &        5.8 &      -18.1 &       13.6 &      -25.0 &       11.0 &      -50.9 \\
\bottomrule
\end{tabular}
}
  \caption{
Results of evaluating DEP, POS, and NER using the seven contextual LMs (random initialized) for Cross-Topic and the difference to their pre-trained counterparts in \autoref{tab:exp1}.} 
\label{tab:random-cross}
\end{table}

\subsection{The Effect of Removing Random Information}
\label{sec:appendix-random-information}

\begin{figure*}[]
  \centering
  \includegraphics[width=1.0\textwidth]{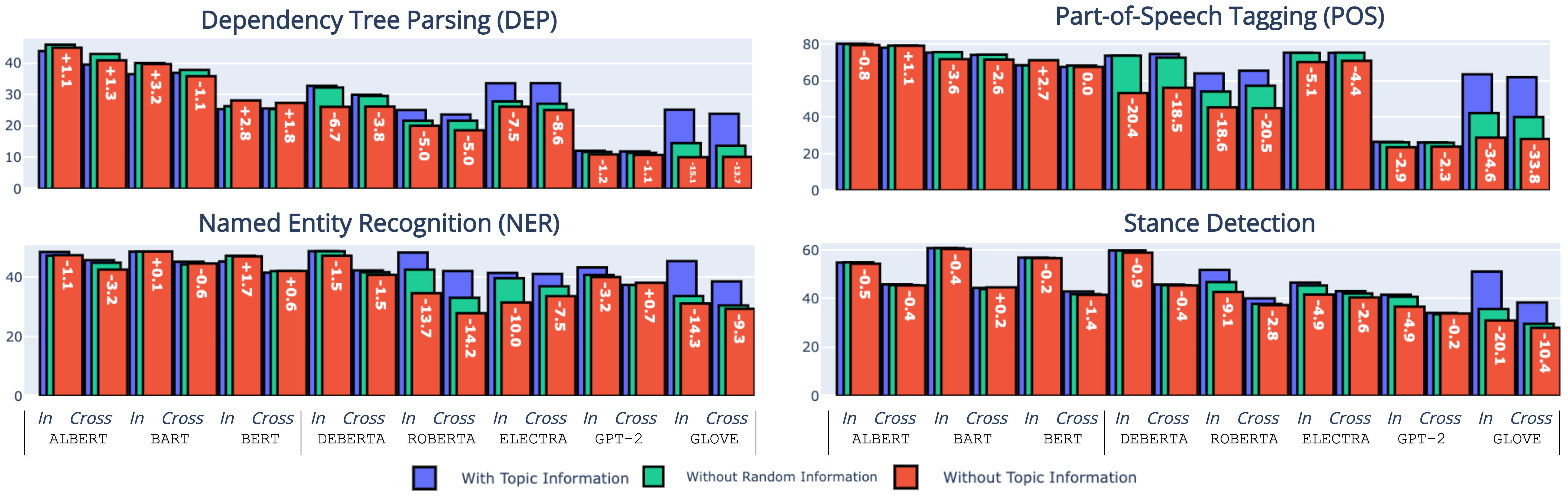}
  \caption{Comparison of the probing results with (blue bars) and without (red bars) topic information, or without random information (green bars). The white text indicates the difference between the blue and red bars. }
  \label{fig:diff-rand}
\end{figure*}

We saw in \autoref{sec:experiment-2} that removing topic-specificity has a big impact for some models (like \ROBERTA~or \ELECTRA) but at the same time can even boost the performance of others like \BERT. 
As suggested in \newcite{elazar-etal-2021-amnesic}, we apply a sanity check by removing random information from the encodings of LMs. 
Following the results in \autoref{fig:diff-rand}, removing random information (green bars) performs in between the scenarios with (blue bars) or without (red bars) topic information for cases where we see a clear negative effect when removing topic information. 
In contrast, removing random information can produce a more pronounced effect when we see performance improvements. 
This observation backs our assumption that removing information can have a regularization effect.

\subsection{The Effect of Removing Topic Information on \textit{Seen} and \textit{Unseen} Instances}
\label{sec:appendix-amnesic-seen-unseen}
 
\begin{figure}[]
  \centering
  \includegraphics[width=0.48\textwidth]{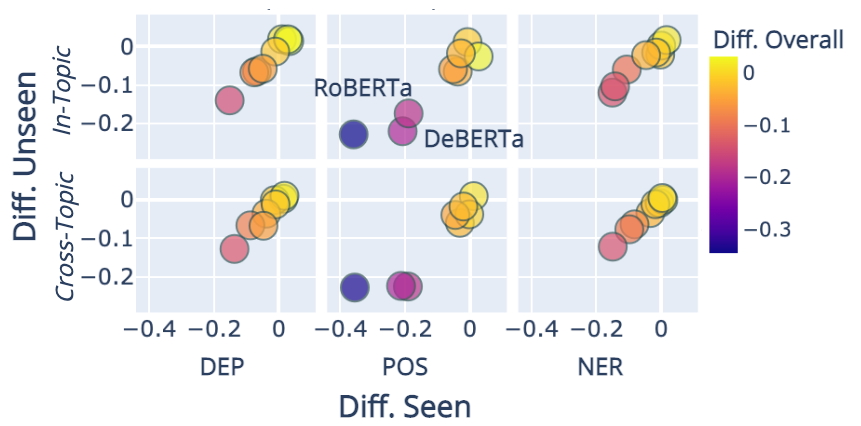}
  \caption{Performance difference for \textit{seen} (x-axis) and \textit{unseen} (y-axis) instances when removing topic information or not.
  One dot represents one LM.}
  \label{fig:amnesic-seen-unseen}
\end{figure}

We show in \autoref{fig:amnesic-seen-unseen} that a performance drop affects \textit{seen} and \textit{unseen} instances for In- and Cross-Topic equally.
Exceptionally, we see \textit{unseen} ones are more affected on POS for \DEBERTA~and \ROBERTA. 
This result indicates that these LMs fall short of generalizing towards rare vocabularies - like \textit{unseen} instances of POS. 

\subsection{Analysis of Per-Class Results for NER}
\label{sec:appendix-ner}

When considering the per-class results of NER in \autoref{tab:exp1-classes}, we see the classes CARDINAL, MONEY, ORG, and PERSON show the biggest differences between In- and Cross-Topic. 
For ORG and PERSON, we see their topic-specific terms as the main reason for the performance gap. 
In contrast, we were surprised about the high difference for CARDINAL.
We think this is mainly because this class embodies all numbers belonging to no other class. 
For MONEY, we see its uneven distribution over topics as the main reason for the performance difference - one topic covers more than 50\% of the instances.
These entities are highly topic-specific from a statistical point of view.

Despite having almost the same performance for In-Topic, \BART~and \DEBERTA~tend to outperform \ALBERT~on classes with more semantic complexities - like GPE, ORG or PERSON.
For Cross-Topic, we see \ALBERT~performing better in classes unevenly distributed instances over topics - like MONEY. 
Further, it outperforms \BART~and \DEBERTA~on less semantical classes (CARDINAL, ORDINAL, PERCENT).

\begin{table}[]
\centering
    \resizebox{0.48\textwidth}{!}{%
  \setlength{\tabcolsep}{3pt}
\begin{tabular}{llccccccccc}
\toprule
   &  &  CARDINAL  &  DATE  &  GPE  &  MONEY  &  NORP  &  ORDINAL  &  ORG  &  PERCENT  &  PERSON  \\ \midrule

 \multirow{3}{*}{\rotatebox[origin=c]{90}{\textit{In}}}  & \ALBERT &       95.0 &   95.3 &  89.4 &    95.0 &   91.3 &      97.8 &  80.2 &      99.2 &     82.7 \\
      & \BART &       94.8 &   94.6 &  89.7 &    95.6 &   91.6 &      97.3 &  81.0 &      99.4 &     83.5 \\
      & \DEBERTA &       95.3 &   95.6 &  90.0 &    96.5 &   91.5 &      97.4 &  81.1 &      99.2 &     83.7 \\ \midrule
 \multirow{3}{*}{\rotatebox[origin=c]{90}{\textit{Cross}}}  & \ALBERT &       91.2 &   95.0 &  88.6 &    55.6 &   90.8 &      98.1 &  78.8 &      98.9 &     81.7 \\
      & \BART &       90.1 &   94.2 &  88.9 &    35.0 &   90.7 &      97.6 &  79.1 &      98.8 &     81.8 \\
      & \DEBERTA &       88.3 &   95.3 &  88.6 &     0.0 &   90.5 &      97.5 &  79.8 &      98.6 &     81.8 \\

\bottomrule
\end{tabular}
}
  \caption{Per-class results of \ALBERT, \BART, and \DEBERTA~on NER for In- and Cross-Topic.
  }
\label{tab:exp1-classes}
\end{table}

\subsection{Effect of Removing Token-Level Topic Information of Per-Class Results for NER}
\label{sec:appendix-ner-without}

Similar to the previous analysis, there are apparent effects of removing topic information when considering NER classes separately.
\autoref{tab:exp2-classes} shows these results for \BART, \BERT, \DEBERTA, and \ROBERTA.
Like the overall result, \BART, \DEBERTA, and \ROBERTA~perform less when removing topic information.
Whereby the effect is the most pronounced for \ROBERTA~  with the highest performance drop for In- and Cross-Topic on classes like NORP or ORDINAL. 
In addition, these results show that the performance gain from removing topic information within \BERT~happens on MONEY for In-Topic and NORP for Cross-Topic.

\begin{table}[]
\centering
    \resizebox{0.5\textwidth}{!}{%
  \setlength{\tabcolsep}{3pt}
\begin{tabular}{llccccccccc}
\toprule
   &  &  CARDINAL  &  DATE  &  GPE  &  MONEY  &  NORP  &  ORDINAL  &  ORG  &  PERCENT  &  PERSON  \\ \midrule
 \multirow{3}{*}{\rotatebox[origin=c]{90}{\textit{In}}} & BART &      -0.23 &   0.04 &  0.15 &    0.15 &   0.02 &     -0.04 &  0.08 &     -0.13 &     0.20 \\
   & BERT &       1.65 &  -0.15 & -0.04 &   28.00 &  -0.14 &     -0.58 &  0.06 &      0.00 &     0.22 \\
   & DEBERTA &      -1.14 &  -0.13 & -1.48 &   -7.74 & -14.40 &     -0.30 & -0.82 &     -0.12 &    -0.10 \\
   & ROBERTA &      -6.00 &  -3.00 & -7.82 &  -24.09 & -90.61 &    -98.06 & -2.66 &     -0.51 &    -0.58 \\\midrule

 \multirow{3}{*}{\rotatebox[origin=c]{90}{\textit{Cross}}} & BART &      -0.48 &   0.01 & -0.13 &    2.45 &  -0.06 &     -0.52 & -0.38 &     -0.09 &    -0.03 \\
   & BERT &      -0.05 &  -0.05 &  1.00 &    0.00 &   8.95 &     -0.60 &  0.29 &      0.00 &     0.00 \\
   & DEBERTA &      -0.07 &  -0.16 & -2.52 &    0.00 & -21.88 &     -0.35 & -0.91 &     -0.01 &     0.07 \\
   & ROBERTA &      -9.04 &  -2.63 & -7.45 &    0.00 & -85.23 &    -98.07 & -2.99 &    -35.97 &    -0.46 \\
\bottomrule
\end{tabular}
}
  \caption{Class-wise effect on the performance when removing topic information of \BART, \BERT, \DEBERTA,  and \ROBERTA~on NER for In- and Cross-Topic.
  }
\label{tab:exp2-classes}
\end{table}

\subsection{The Effect of Fine-Tuning on NER Classes}\label{sec:appendix-ner-ft}

Analysing the results (\autoref{tab:exp3-classes}) for every NER class gives additional insights into where the fine-tuning had the most significant effect.
We generally see the biggest effect on classes with less semantic meaning, like ORDINAL, PERCENT, or MONEY.
At the same time, GPE, PERSON, and ORG are less affected as classes with more attached semantics.
Regarding the different LMs, \ALBERT~and \DEBERTA~show the most performance training, while \BERT~gains performance for the MONEY class. 

\begin{table}[]
\centering
    \resizebox{0.5\textwidth}{!}{%
  \setlength{\tabcolsep}{3pt}
\begin{tabular}{llccccccccc}
\toprule
   &  &  CARDINAL  &  DATE  &  GPE  &  MONEY  &  NORP  &  ORDINAL  &  ORG  &  PERCENT  &  PERSON  \\ \midrule
 \multirow{4}{*}{\rotatebox[origin=c]{90}{\textit{In}}} & ALBERT &      -34.2 &  -25.4 & -26.9 &   -95.0 &  -51.9 &     -60.3 & -22.4 &     -99.2 &    -21.8 \\
   & BART &       -8.5 &   -7.2 &  -7.5 &    -7.2 &  -10.4 &     -36.6 &  -4.1 &      -3.8 &     -2.7 \\
   & BERT &       -1.9 &   -2.0 &  -2.0 &    34.8 &   -4.4 &     -17.9 &  -0.8 &      -3.9 &     -1.1 \\
   & DEBERTA &      -15.1 &   -6.8 &  -8.7 &   -19.5 &  -43.7 &     -60.8 &  -8.8 &     -24.8 &     -8.3 \\\midrule

 \multirow{4}{*}{\rotatebox[origin=c]{90}{\textit{Cross}}} & ALBERT &      -21.5 &  -10.4 & -19.1 &   -55.6 &  -34.4 &     -13.1 & -10.7 &     -81.0 &     -9.2 \\
   & BART &       -9.2 &   -7.4 &  -7.0 &   -16.3 &  -11.2 &     -24.4 &  -3.9 &      -4.5 &     -2.1 \\
   & BERT &       -2.5 &   -1.2 &  -1.2 &     3.6 &   -2.2 &      -9.7 &  -0.8 &      -2.6 &     -0.5 \\
   & DEBERTA &      -18.2 &   -6.2 & -12.7 &     0.0 &  -50.6 &     -76.0 & -11.7 &     -73.5 &     -6.8 \\
\bottomrule
\end{tabular}
}
  \caption{Per-class difference before and after fine-tuning on \emph{stance detection} of \ALBERT, \BART, \BERT, and \DEBERTA~on NER for In- and Cross-Topic.
  }
\end{table}
\label{tab:exp3-classes}

\begin{table}[h]
\centering
    \resizebox{0.48\textwidth}{!}{%
  \setlength{\tabcolsep}{2pt}
\begin{tabular}{l@{\htc3}cc@{\htc3}cc@{\htc3}cc@{\htc3}cc@{\htc3}:ccc}
\toprule
 & \multicolumn{2}{c@{\htc3}}{\textbf{DEP}} & \multicolumn{2}{c@{\htc3}}{\textbf{POS}} & \multicolumn{2}{c@{\htc3}}{\textbf{NER}}\\\midrule
 & \textit{In} & \textit{Cross}&  \textit{In} & \textit{Cross}&  \textit{In} & \textit{Cross}\\
\midrule
\ALBERT & 85.2 & 83.9 & 93.8 & 93.6 & 86.9 & 85.0 \\
\BART & 80.9 & 81.0 & 92.6 & 92.0 & 87.1 & 84.5 \\
\BERT & 76.1 & 76.1 & 89.2 & 88.6 & 85.2 & 82.9 \\
\DEBERTA & 81.2 & 79.9 & 92.8 & 93.1 & 87.5 & 84.0\\
\ROBERTA & 75.9 & 75.5 & 89.6 & 90.1 & 86.3 & 83.2 \\
\ELECTRA & 81.1 & 80.7 & 92.3 & 92.2 & 82.8 & 82.2 \\
\GPT & 69.8 & 69.1 & 85.8 & 85.7 & 84.6 & 81.1\\
\GLOVE & 39.5 & 38.5 & 46.6 & 45.9 & 78.8 & 77.2 \\
\hdashline
\textit{\textbf{Average}}  & 73.7 & 73.1 & 85.3 & 85.2 & 84.9 & 82.5\\ 
\midrule
\BERT~80k & 80.5 & 79.1 & 92.0 & 91.5 & - & - \\
\BERT~160k & 84.3 & 84.2 & 93.1 & 92.8 & - & - \\
\BERT~320k & 86.3 & 85.6 & 93.7 & 93.3 & - & - \\
\midrule
\BERT~\citep{Tenney2019WhatDY}  &  \multicolumn{2}{c@{\htc3}}{93.0}  & \multicolumn{2}{c@{\htc3}}{97.0}  & \multicolumn{2}{c@{\htc3}}{96.1} \\
\BERT~\citep{tenney-etal-2019-bert} & \multicolumn{2}{c@{\htc3}}{95.2}  & \multicolumn{2}{c@{\htc3}}{96.5}  & \multicolumn{2}{c@{\htc3}}{96.0} \\
\BERT~\citep{hewitt-liang-2019-designing} &\multicolumn{2}{c@{\htc3}}{89.0}  & \multicolumn{2}{c@{\htc3}}{97.2}  & \multicolumn{2}{c@{\htc3}}{-} \\
\bottomrule
\end{tabular}
}
  \caption{
Accuracy results for In- and Cross-Topic probing results for eight LMs, across three random seeds. Further, we report results of gradually increasing the number of consider instance (\BERT~80k, \BERT~160k, and \BERT~320k), as well as reference performance of previous work \citep{Tenney2019WhatDY,tenney-etal-2019-bert,hewitt-liang-2019-designing}.
} 
\label{tab:exp1-acc}
\end{table}

\subsection{Annotation Verification}\label{sec:appendix-label-verification}

To evaluate probing tasks in the In- and Cross-Topic setup, we rely on data with topic annotations on the instance level - like the \textit{UKP ArgMin} \citep{stab-etal-2018-cross} or the \textit{wtwt} \citep{conforti-etal-2020-will} dataset.
Since these datasets do not include linguistic annotations, we make use of spaCy\footnote{https://spacy.io/} to automatically derive the labels for \emph{dependency tree parsing (DEP)}, \emph{part-of-speech tagging (POS)}, or \emph{named entity recognition (NER)}.
We used the \texttt{en\_core\_web\_sm} model, which provides reliable labels with a detection performance in terms of accuracy of 97.0 for POS, 90.0-92.0 for DEP, and an F1 score of 85.0 for NER (details available \href{https://spacy.io/models/en}{online}).
Note, this performance referees to identify valid candidates (like entities for NER) given a piece of text, and assign the corresponding labels, such as person or organization.
In contrast, in probing, we consider only the second step: assigning the right label of a valid candidate. 
Therefore, we can not directly compare recognition and probing performance. 

Considering our results (\autoref{sec:experiment-1}), we see these derived labels as reliable and well aligned with previous work \citep{Tenney2019WhatDY,tenney-etal-2019-bert,hewitt-liang-2019-designing}, even though we mainly report $F_1$ score.
One reason for that is the similar performance ranking (DEP $<$ NER $<$ POS) as in previous work, considering $F_1$ score as well as the accuracy score reported in \autoref{tab:exp1-acc}. 
Another reason is the narrowing accuracy performance gap between our experiments and previous work when we gradually increase the number of consider instance from 40k to 80k, 160k, until 320k.

\end{document}